\documentclass[12 pt, letterpaper]{article}

\usepackage{graphicx}
\usepackage{caption}
\usepackage{subcaption}
\usepackage{algorithm}
\usepackage[noend]{algpseudocode}
\usepackage{indentfirst}

\title{Risk Factor Identification In Osteoporosis Using Unsupervised Machine Learning Techniques}
\author{Mikayla Calitis \\
	Department of Computer Science \\
	Misericordia University}
\date{December 15, 2023}

\begin{document}
	\maketitle
	
	\begin{abstract}
		In this study, the reliability of identified risk factors associated with osteoporosis is investigated using a new clustering-based method on electronic medical records. This study proposes utilizing a new CLustering Iterations Framework (CLIF) that includes an iterative clustering framework that can adapt any of the following three components: clustering, feature selection, and principal feature identification. The study proposes using Wasserstein distance to identify principal features, borrowing concepts from the optimal transport theory. The study also suggests using a combination of ANOVA and ablation tests to select influential features from a data set. Some risk factors presented in existing works are endorsed by our identified significant clusters, while the reliability of some other risk factors is weakened.		
	\end{abstract}
	
	\newpage
	\section{Introduction}		
	The National Institute of Health (NIH) defines osteoporosis as a bone disease characterized by decreased bone density and strength, leading to an increased risk of bone fractures and reduced quality of life from losses in function. Osteoporosis is expensive, costing the healthcare system \$19 billion yearly. Currently, osteoporosis does not have a cure or treatment and requires an early diagnosis for prevention; however, osteoporosis is a ``silent disease" and does not typically have symptoms until after a fracture occurs \cite{wang2006hybrid, turkyosteoporosis, nof2015}. Some previously established risk factors include gender, age, ethnicity, physical activity levels, weight, familial history, and occupation. Identifying risk factors of osteoporosis is critical in diagnosing the disease in its early stages to implement preventative treatment. A great number of risk factors were proposed in a variety of previous works; however, the reliability of these risk factors has not been well investigated. To study the reliability, more scientific methods are needed. Thus, we propose utilizing a novel CLustering Iterations Framework (CLIF), Wasserstein distance metrics in principal feature identification, and combining ablation and ANOVA during feature selection. 
	
	CLIF iteratively identifies dense clusters in which data points in the cluster are concentrated and close together, removes them, and re-clusters the remaining data to find dense clusters and sub-clusters for analysis. It includes an iterative clustering framework that can adapt any clustering method, a feature selection component, and a principal feature identification component. Wasserstein distance borrows concepts from the optimal transport theory to identify principal features of clusters. The study also proposes using a combination of ANOVA and ablation tests that remove parts of the data set to observe changes in performance outcomes and assist in selecting influential features from a data set. Using these proposed methods, this study found multiple dense clusters with a density of 0.85 or greater in five iterations of Hierarchical Density-Based Spatial Clustering of Applications with Noise (HDBSCAN), separating them by the following principal features: age, history of a wrist fracture, no history of a wrist fracture, no history of hip fracture, no history of a spinal fracture, daily prednisone or cortisone use, mother was diagnosed with osteoporosis, and father was diagnosed with osteoporosis. 
	
	Section 2 of this paper discusses previous research related to osteoporosis studies and data mining studies. Section 3 explains the proposed CLIF for clustering, Wasserstein distance metrics for principal feature analysis, and a combination of ablation algorithms and ANOVA testing for feature selection. Section 4 explains the implementation of the suggested techniques, including the data collection, methodology, and results. Section 5 discusses the limitations of the study, and section 6 highlights the study, methods, and findings. Section 7 proposes future studies needed to gain a better understanding of the risk factors associated with osteoporosis.

	\section{Related Works}
	Presently known risk factors of osteoporosis include low peak bone mass in adolescence, lactation duration in women, history of surgery, number of pregnancies in women, removal of the ovaries at an early age in women, prolonged immobility, and prolonged use of corticosteroids \cite{pouresmaeili2018comprehensive, saggese2001osteoporosis, jabarpour2020osteoporosis, kelsey1989risk, kanis1998risk}. Immune-suppressants, hyperlipidemia drugs, anti-inflammatory drugs, anti-coagulant drugs, and hypertension drugs may also increase the risk of osteoporosis \cite{jabarpour2020osteoporosis}. Some factors are not modifiable, such as fall history, old age, gender, white ethnicity, prior fractures, and family history. A few adjustable risk factors are nutritional absorption, vitamin D intake, calcium intake, physical activity, fall risk, weight, cigarette smoking, alcohol consumption, and stress \cite{pouresmaeili2018comprehensive}. Pouresmaeili et al. \cite{pouresmaeili2018comprehensive} also identified comorbidities, hypogonadism, hyperparathyroidism, hyperlipidemia, thalassemia, liver disease, inflammatory diseases, renal disease, cardiovascular disease, diabetes mellitus, and dementia, correlated with osteoporosis risk factors. Multiple studies, such as the ones referenced above, address the problem of osteoporosis predictions.
	
	EHRs and medical records have a lot of data on a patient, including a patient's physical parameters, lifestyle, laboratory values, case history, and treatment modality. Mining medical data can reveal class or concept descriptions of medical terms, search for association rules, classify patients and predict medical events of new patients, search for clusters from different points of view, and carry out evolution analysis based on time-series data \cite{vathyapplication}. One study discusses data mining techniques in a medical domain, considering that EHRs have dirty data that can be redundant, have missing values, and unstructured texts \cite {vathyapplication}. Other data mining studies used several surveys and health records, exhibiting potential for osteoporosis predictions. A few of these data sets are from the clinical history of the osteoporosis research center of Tehran University of Medical Sciences between 2006 and 2010 \cite{jabarpour2020osteoporosis}, the Rochester Epidemiology Project (REP) medical records linkage system \cite{wang2020unsupervised}, and a cross-sectional study conducted at the Universidade do Extremo Sul Catarinense (UNESC) \cite{de2016use}. Additional researchers applied data mining techniques to hospital questionnaires, diagnosis results, and medical images \cite{wang2006hybrid, lu2020research}. 
	
	Several studies implemented supervised machine learning models in previous methodologies. Wang et al. \cite{wang2006hybrid} used neural networks and decision trees to implement hybrid ensembles in their research and found a relatively high level of diversity and improvement in prediction accuracy. Saggese et al. found that Support Vector Machine (SVM) and Bayes methods (Tree Augmented Naïve Bayes (TAN)) had 88.39\% and 91.29\% classification precision rates, respectively \cite{jabarpour2020osteoporosis}. Supervised learning may not be optimal for disease prediction models because they require human input, labeling, and training that can take extensive amounts of time \cite{book-healthcareanalytics}. Additionally, neural network decisions are not easily understood and interpreted, and therefore difficult to apply the findings to the healthcare field \cite{book-healthcareanalytics, book-modelbasedcluster}. Neural networks inadequately perform when handling missing data, high amounts of noise, and overlaps, which is common in medical records \cite{book-healthcareanalytics}. 
	
	Studies were also conducted with unsupervised machine-learning techniques. Wang et al. \cite{wang2020unsupervised} concluded that Poisson Dirichlet Model (PDM) can identify disease clusters based on latent patterns and that the Latent Dirichlet Allocation (LDA) model can stratify patients into subgroups. Clustering methods are advantageous over topic models because they separate groups into subgroups with the expected content linked by a theme of common attributes, unlike topic models that retrieve keywords for a topic \cite{book-datacluster}. Hierarchical clustering includes independently obtaining the optimal number of clusters, providing the most similar observations to given observations, being easily adapted to categorical variables, and being less sensitive to outliers. Kruse et al. \cite{kruse2017clinical} identified four clusters for low bone mineral density (BMD) and two clusters for young age with high BMD. One study used a fuzzy c-means analysis in which each data point belongs to all of the clusters by a differing probability and found eight multi-morbidity clusters that are possible risk factors of osteoporosis, present in 93.1\% of the sample possible \cite{violan2019soft}.
	
	\section{Methods}
	This study proposes CLIF and Principal Feature Identification using Wasserstein distance metrics to mine relationships between selected risk factors and osteoporosis in participants of the Continuous NHANES data \cite{NHANES}, ranging from years 1999 to 2020.
	
	The first proposition of this study is the application of CLIF to a clustering method, a feature selection method, and a principal feature identification method. Iteratively identifying and removing dense clusters reveals subtle differences in cluster groups. Dense clusters found in a single iteration can have prominent features that mask subtle differences that may be present in the subset of the data. By removing the top clusters, patterns may be visible in dense sub-clusters. CLIF is implemented in this study because it is a framework applicable to any clustering algorithm. CLIF identifies the most dense clusters in the data set sorted by another clustering algorithm. The data points in these clusters are removed, and new dense clusters are found in the remaining data. The dense clusters are extracted in each iteration until there are no dense clusters left to be identified. To determine the discriminating features of the clusters, the principal feature identification method is performed for each iteration and each pair of clusters in that iteration (See Algorithm \ref{alg:CLIF}).
	\begin{algorithm}
		\caption{CLIF}
		\label{alg:CLIF}
		\begin{algorithmic}
			\Function{Clustering}{dataset}
				\State Cluster data points
				\State Retrieve density values
				\State Remove clusters with density $>$ density threshold
				\State 
				\Return clusters and iteration number
			\EndFunction
			\Function{Principal Feature Identification}{clusters}
				\ForAll {Selected clusters}
					\ForAll {Features}
						\State Compare feature 1 in cluster A and feature 1 in cluster B
						\If {The score is significantly different}
							\State
							\Return Principal feature
						\EndIf
					\EndFor
				\EndFor
			\EndFunction
			\Function{Feature Selection}{dataset}
				\State Measure feature importance
				\State
				\Return Top features
			\EndFunction
			\\
			\Call{Feature Selection}{}
			\State
			\Call{Clustering}{}
			\State
			\Call{Principal Feature Identification}{}
			\While {number of dense clusters $>$ 0}
			\State
			\Call{Clustering}{}
			\State
			\Call{Principal Feature Identification}{}
			\EndWhile
		\end{algorithmic}
	\end{algorithm}
	
	The following proposition of this study is to use the Wasserstein distance metric to identify the principal features of each cluster. At each iteration, the dissimilarities of the data points, represented by a vector of feature values, separate the identified dense clusters. This distance between clusters results in a significant difference in values of one or multiple features. In other words, each cluster is distinguishable by at least one of its features (See Algorithm \ref{alg:PFI}) \cite{Fanchao}. To observe differences in features, the Wasserstein distance metric compares the feature distribution in each pair of clusters. The Wasserstein distance uses optimal transport to compare distributions by the minimum cost required to transform one distribution into another. The cost is calculated with the amount of data that is moving and the distance traveled. Advantages of Wasserstein distance include stability with overlapping and multimodal distributions, the ability to compare differently sized and shaped structures by considering distances between points, and its robustness to noise providing more reliable dissimilarity measures \cite{Rahman_2023} (See Algorithm \ref{alg:Wasserstein}).
	\begin{algorithm}
		\caption{Principal Feature Identification}
		\label{alg:PFI}
		\begin{algorithmic}
			\For{$x$ in [A, B, C]}
				\For{$y$ in [A, B, C]}
					\If{$x==y$:}
						\State continue
					\Else
						\For{att in [Att1, Att2, Att3]}
							\State d(att, x, y) = Distance between the distribution of Att1 in x and the distribution of Att1 in y
						\EndFor
					\EndIf
				\EndFor
			\EndFor
		\end{algorithmic}
	\end{algorithm}
	
	\begin{algorithm}
		\caption{Principal Feature Identification with Wasserstein Distance Metric}
		\label{alg:Wasserstein}
		\begin{algorithmic}
			\For {d(att, x, y)}
				\State Look at how to transport x on y with transport map T
				\ForAll{T}
					\State Look at total distance traveled
				\EndFor
				\State
				\Return T with minimum (optimal) distance traveled
			\EndFor
			\State 
			\Return Wasserstein Distance for d(att, x, y)
		\end{algorithmic}
	\end{algorithm}
	
	Another proposition of this study is using a combination of feature selection techniques for dimensionality reduction, including ANOVA and ablation. The data is large and contains a mix of numerical and categorical data. To get a better estimate of significant features, ablation and ANOVA tests are conducted separately, and the results are compared. The ANOVA test calculates feature variability and differences between group means. Ablation is an algorithm in which parts of the machine learning model are removed to measure the causality in the change in results. Features that are similarly highly ranked are selected and used in the reduced data set in the rest of the study. For the remaining top features that had discrepancies in results, the ANOVA test was more prominent in determining its significance, considering that a majority of the data is categorical (See Algorithm \ref{alg:AblationAnova}).
	\begin{algorithm}
		\caption{Feature Selection with Ablation and ANOVA}
		\label{alg:AblationAnova}
		\begin{algorithmic}
			\Function{Feature Selection}{dataset}
				\Function{Ablation}{}
					\ForAll{Columns in dataset}
						\State Delete column
						\State 
						\Return Performance
					\EndFor
					\State 
					\Return List of columns sorted by performance score
				\EndFunction
				\If{top ANOVA features match top Ablation features}
					\State Select features to create a decreased dimensionality set
				\Else
					\State Select more features based on the ANOVA score
				\EndIf
			\EndFunction
		\end{algorithmic}
	\end{algorithm}
	
	This study utilizes unsupervised machine learning models to cluster and analyze the dataset. Unsupervised learning techniques allow for pattern discovery for unlabeled data and provide insights without explicit guidance or instruction. They do not need to be trained before finding patterns in the data, and therefore cost less time and can find hidden patterns in the data. Unsupervised machine learning can assist in understanding patterns in EHRs, while supervised machine learning assists in predicting classifications \cite{kruse2018new,wang2020unsupervised}. Clustering methods are advantageous because they separate groups into subgroups with the expected content linked by a theme. Clustering divides the data into clusters, creating more clusters according to common attributes \cite{book-datacluster}. Hierarchical clustering includes independently obtaining the optimal number of clusters, providing the most similar observations to given observations, being easily adapted to categorical variables, and being less sensitive to outliers. HDBSCAN is a clustering algorithm that performs hierarchal clustering over Density-Based Spatial Clustering of Applications with Noise (DBSCAN), in which using various epsilon values allows the algorithm to find clusters of varying densities and is more robust to parameter selection. The algorithm can discover clusters of arbitrary shape, is insensitive to noise and outliers, and does not require prior specification of the number of clusters. EHRs are very likely to have missing and noisy data, and the shape of the clusters is unknown.
	
	The density of the cluster is calculated using a ``k-nearest neighbor search" on the mediods. The clusters produced by HDBSCAN may not be spherical, so a medoid better represents the cluster center than the centroids. Mediods are confined to existing data points, whereas a centroid utilizes a mean that can lie outside the cluster if it is not spherical. K-nearest neighbor is a simple and easy-to-understand implementation ideal for non-linear data since there's no assumption about underlying data. Once the distances of the nearest points are used to calculate the cluster's density, the clusters are sorted descending by cluster density to retrieve the labels of the densest clusters.
	
	\section{Experiments}
	
	\subsection{Data}
\begin{figure}[h]
		\centering
		\includegraphics[width=\linewidth]{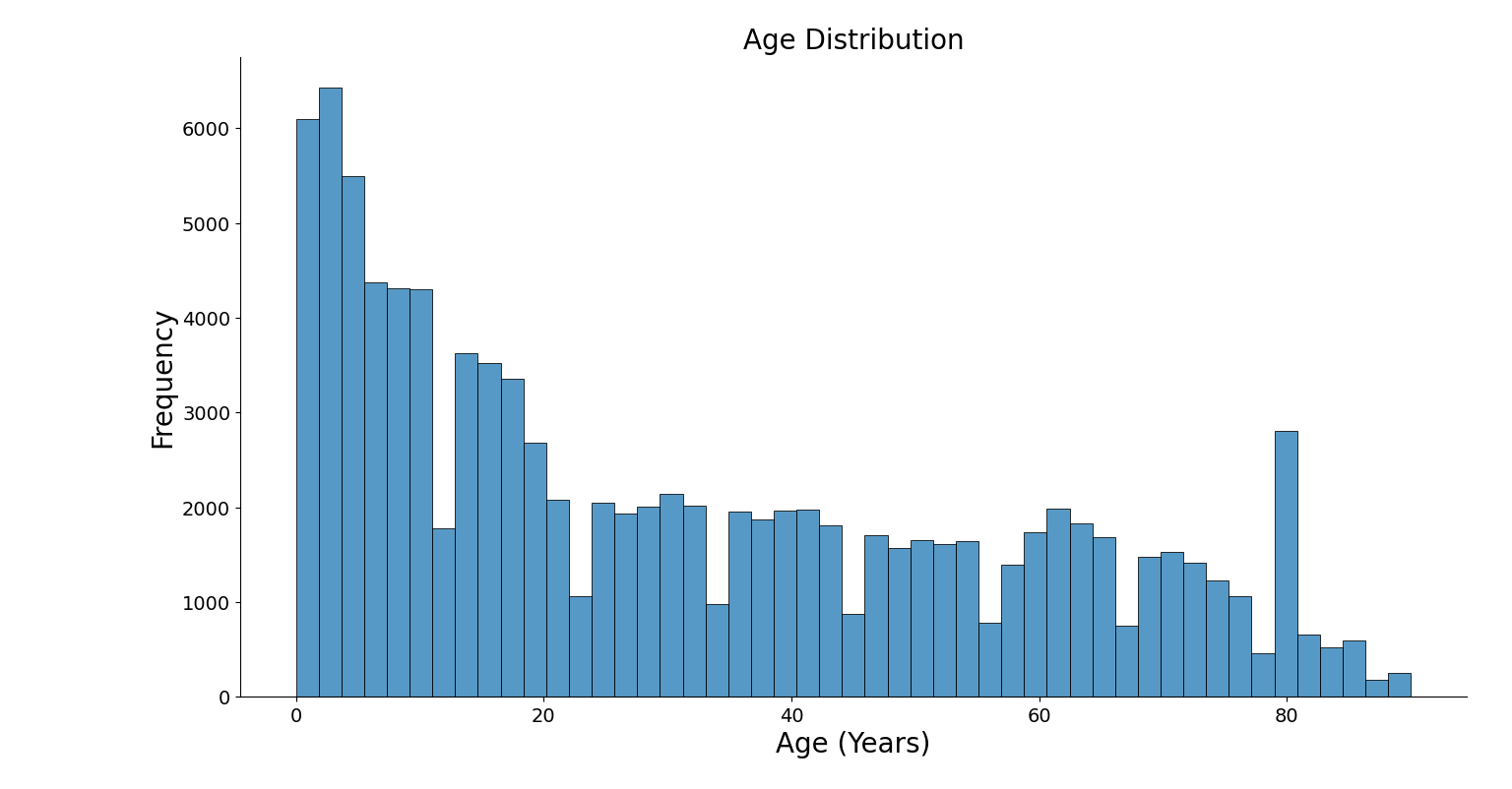}
		\captionsetup{justification=centering, margin=2cm}
		\caption{Distribution of Participant Ages}
		\label{fig:age_dist}
	\end{figure}

\begin{figure}[h]
		\centering
		\includegraphics[width=\linewidth]{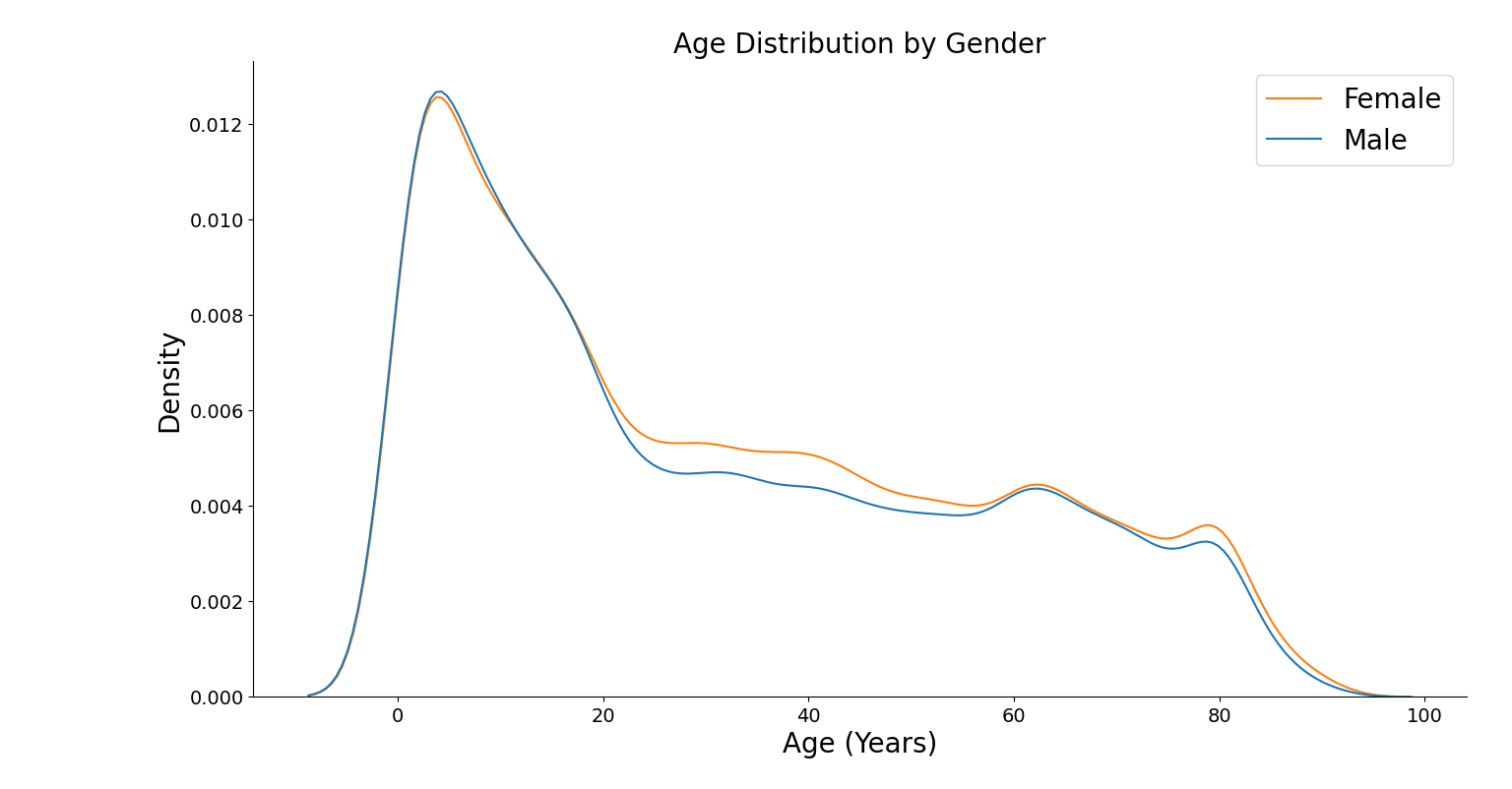}
		\captionsetup{justification=centering, margin=2cm}
		\caption{Distribution of Participant Ages by Gender}
		\label{fig:age_gender}
	\end{figure}

\begin{figure}[h]
		\centering
		\includegraphics[width=0.7\linewidth]{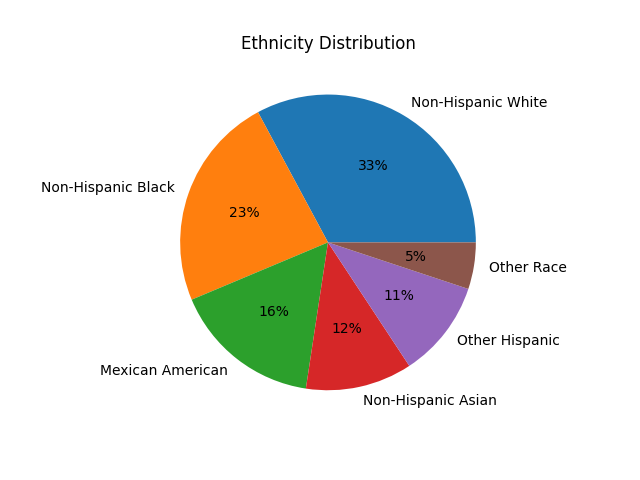}
		\captionsetup{justification=centering}
		\caption{Percentages of Participant Ethnic Background}
		\label{fig:eth_pie}
	\end{figure}

\begin{figure}[h]
		\centering
		\includegraphics[width=\linewidth]{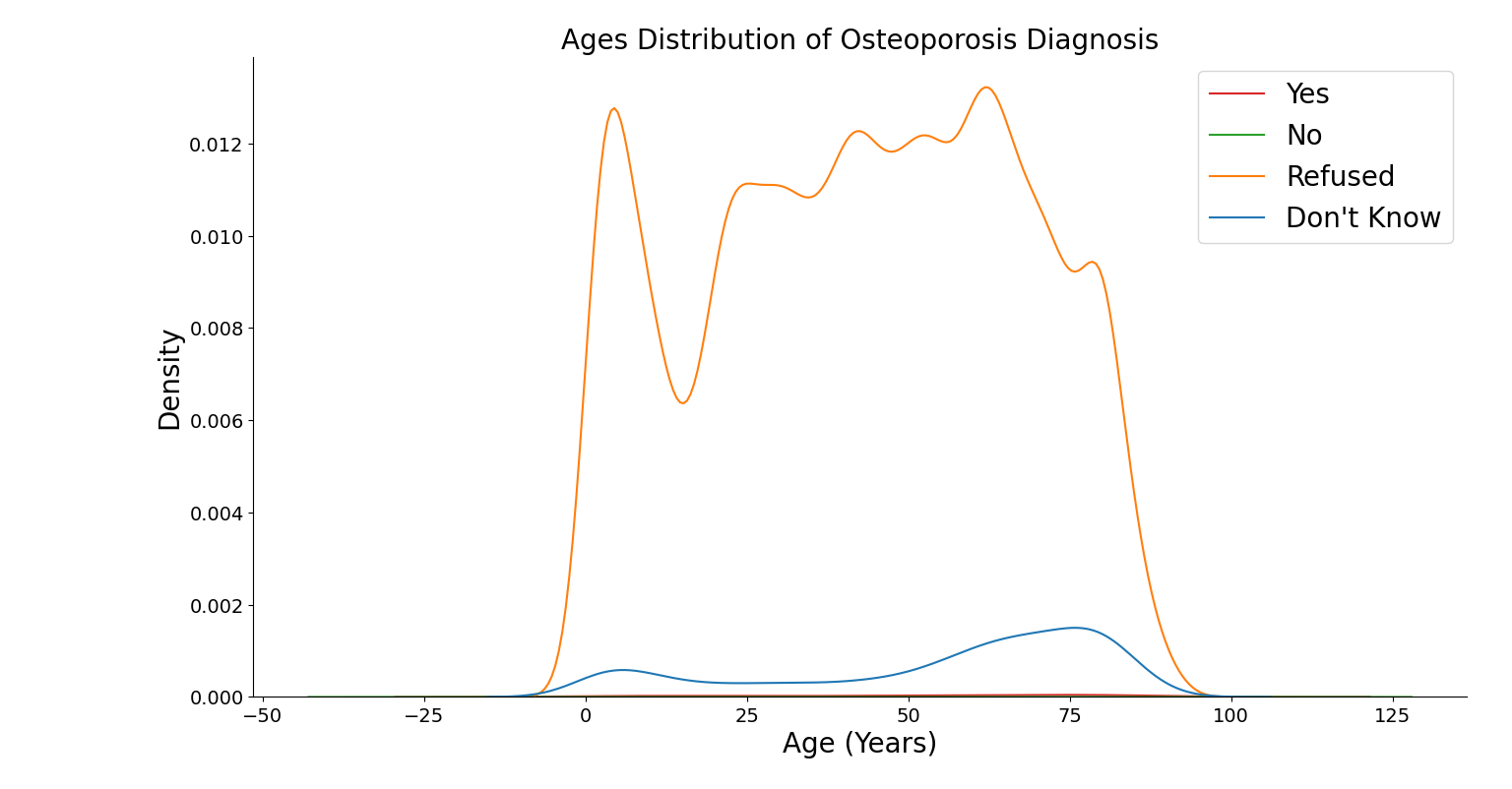}
		\captionsetup{justification=centering, margin=2cm}
		\caption{Age Distribution by Osteoporosis Diagnosis}
		\label{fig:osteo_age}
	\end{figure}

\begin{figure}[!h]
		\centering
		\includegraphics[width=\linewidth]{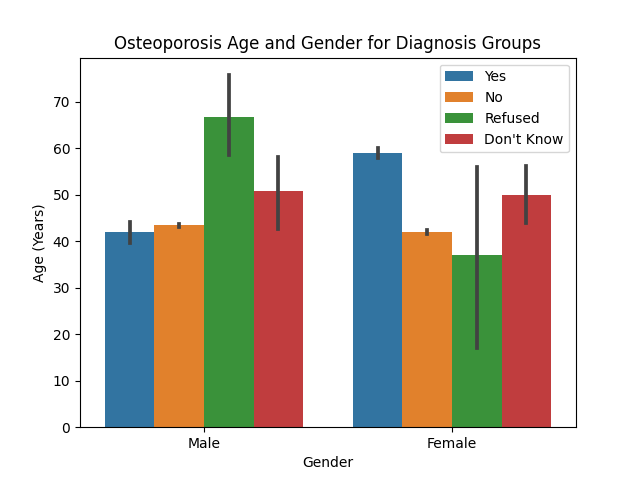}
		\captionsetup{justification=centering, margin=2cm}
		\caption{Average Age of Osteoporosis Patients Grouped by Gender and Diagnosis}
		\label{fig:osteo_age_gender}
	\end{figure}

\begin{figure}[!h]
		\centering
		\includegraphics[width=\linewidth]{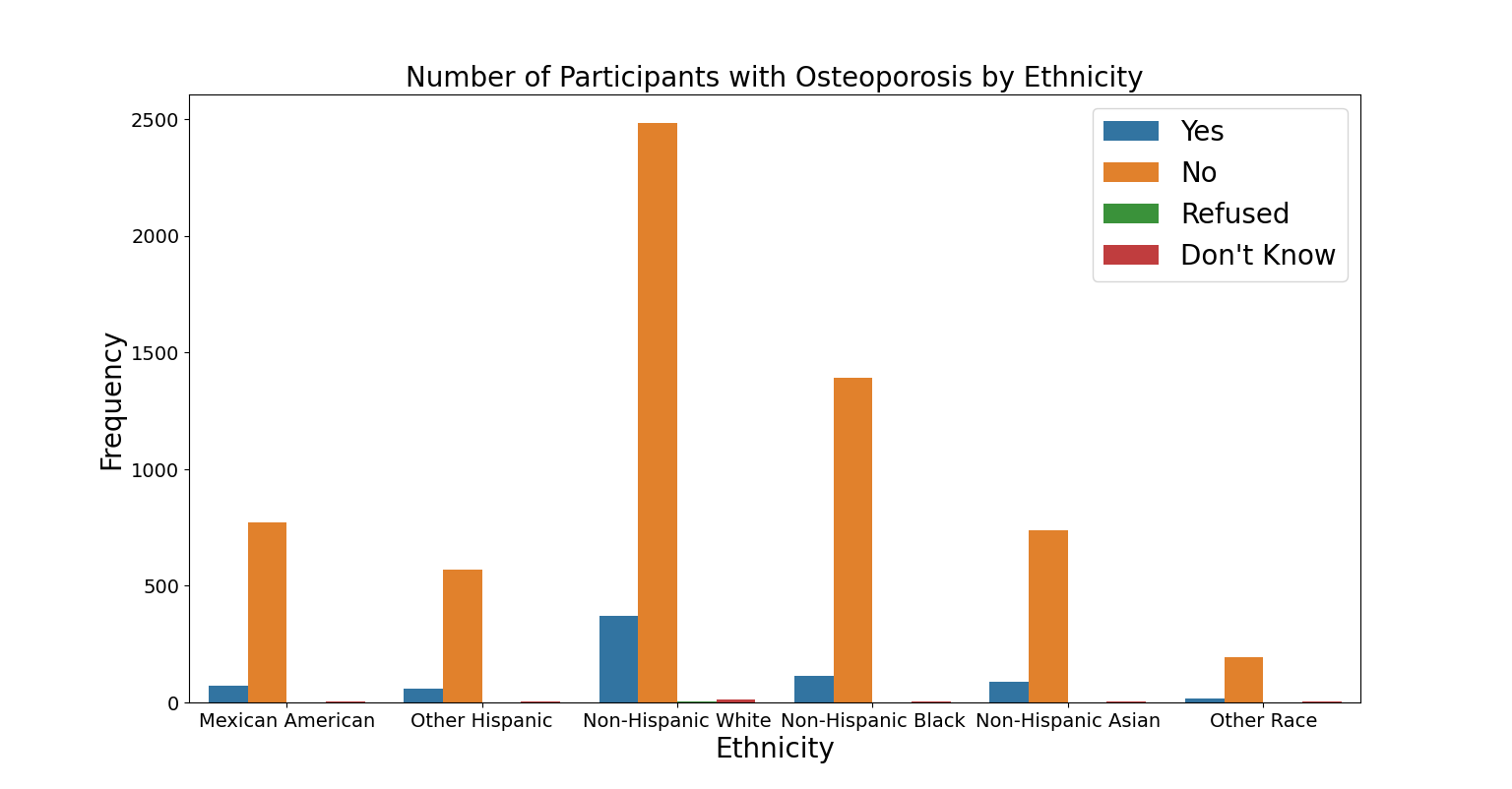}
		\captionsetup{justification=centering, margin=2cm}
		\caption{Osteoporosis Diagnosis for Each Ethnicity}
		\label{fig:osteo_eth}
	\end{figure}

	The data used in the analysis is real data obtained from the National Health and Nutrition Examination Surveys (NHANES) conducted by the National Center for Health Statistics (NCHS) \cite{NHANES}. The NHANES combines data collected from physical examinations and interviews with 5,000 participants each year from 1999 and 2020. The data includes various information on demographics, physical measurements, diet, laboratory test results, health history, and medications. The study will focus on the data related to the listed risk factors. The data has 101,316 participants, ranging between the ages of 0 years old to 90 years old. The ages of the participants are positively skewed, with a mean of 31.31 years old and a standard deviation of 25.26 (See Figure \ref{fig:age_dist}).
	There are 49,516 males and 51,800 females. The distribution for each gender was similar to the total distribution (See Figure \ref{fig:age_gender}).
	There are 6,373 Mexican Americans, 4,164 Other Hispanics, 12,863 Non-Hispanic Whites, 9,194 Non-Hispanic Blacks, 4,566 Non-Hispanic Asians, and 1,996 other races, including multi-racial in the survey and examination data (See Figure \ref{fig:eth_pie}).
	2,598 people answered that a doctor told them that they had osteoporosis, 36,617 answered that a doctor did not tell them that they had osteoporosis, 8 people refused to answer, and 122 people did not know if a doctor told them if they had osteoporosis. The average age of those with osteoporosis identified by a doctor is 54.61 years old, with a standard deviation of 25.93 (See Figure \ref{fig:osteo_age}).
	The minimum age for someone with osteoporosis was 1 year old, and the oldest person was 90 years old. Of the 2,598 people who answered they had been diagnosed with osteoporosis, 666 were male, and 1,932 were female. The average age of males with osteoporosis is 41.92 years old with a standard deviation of 28.77, while the average age of females with osteoporosis is 58.99 years old with a standard deviation of 23.33 (See Figure \ref{fig:osteo_age_gender}).
	Of those that answered yes to being told by a doctor that they have osteoporosis, 71 were Mexican Americans, 59 were Other Hispanics, 371 were Non-Hispanic Whites, 113 were Non-Hispanic Blacks, 87 were Non-Hispanic Asians, and 17 were other races, including multi-racial. Of those that answered no to being told about having osteoporosis, 770 were Mexican Americans, 568 were Other Hispanics, 2,483 were Non-Hispanic Whites, 1,390 were Non-Hispanic Blacks, 736 were Non-Hispanic Asians, and 192 were Other races including multi-racial (See Figure \ref{fig:osteo_eth}).
	
	\subsection{Experimental Methods}
	\subsubsection{Data Collection}
	All of the XPT files on the ``All Continuous NHANES" website were downloaded onto the local machine. The XPT files were then converted to CSV files for usability with the pandas data analysis library and imported into a PostgreSQL database for scalability. The CSV files were merged by category: demographics, dietary, examination, laboratory, and questionnaire. A csvsql tool created a ``create table" SQL statement, and the result was manually modified. The csvsql tool guesses column types as ``BOOLEAN" by default if it cannot make a guess or does not read any valid data for a column. The ``BOOLEAN" type is not always correct, so ``BOOLEAN" types were manually changed to ``VARCHAR" in the generated SQL expression. The other columns are numerical and labeled as ``DECIMAL" types.  Additionally, Postgres has a maximum of 1600 columns per table, the data was split into several tables, using another feature of the csvkit tool, ensuring that headers are well written and that the ``SEQN" which is the participant ID is in each resulting CSV file. The csvsql tool created a ``create table" statement for PostgreSQL, then the COPY command loaded the data from the split CSV files.
	
	\subsubsection{Data Preprocessing}
\begin{figure}[!h]
		\centering
		\includegraphics[width=\linewidth]{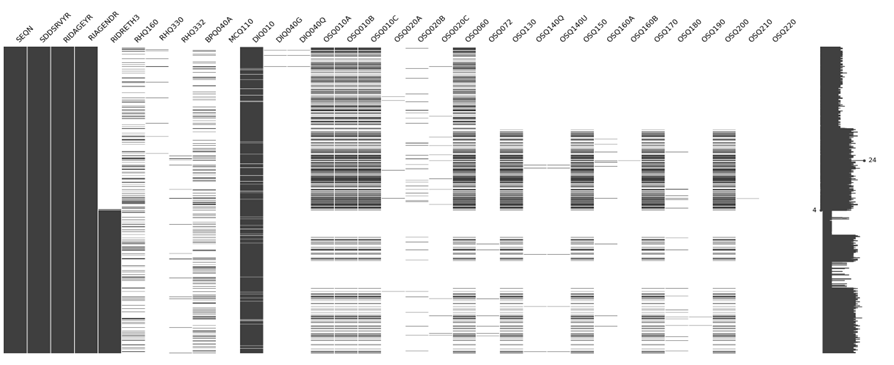}
		\captionsetup{justification=centering, margin=1cm}
		\caption{Matrix of the Missing Data \\ Black areas represent the data that is present. White areas represent the data that is missing.}
		\label{fig:missing_matrix}
	\end{figure}

	32 columns were manually selected based on findings in previous research to reduce the dimensionality of the data, including features related to the subjects’ cohort group, age, gender, ethnicity, pregnancies, ovary removal, hypertension drugs, diabetes, previous hip, wrist, and spinal fractures, osteoporosis diagnosis, daily prednisone or cortisone use, and family history with hip fractures and osteoporosis. The data included many missing values but were not removed due to dependencies on follow-up questions. Those who responded ``no" to certain questions had missing entries for any follow-up questions because they did not apply to the subject(See Figure \ref{fig:missing_matrix}).
	Simple imputation was implemented on the missing data where categorical data was filled with a constant value of ``777" to indicate a missing value consistent with the NHANES coding, and numerical data was imputed with the mean. The data was also one-hot encoded to create new binary columns for each categorical response in the original data column. With 32 columns of data, of which 11 were numerical and 21 were categorical, imputing data and one-hot encoding resulted in 110 features. 
	
	\subsubsection{Application of Methods}
\begin{figure}
		\begin{subfigure}[b]{\textwidth}
			\centering
			\includegraphics[width=\textwidth]{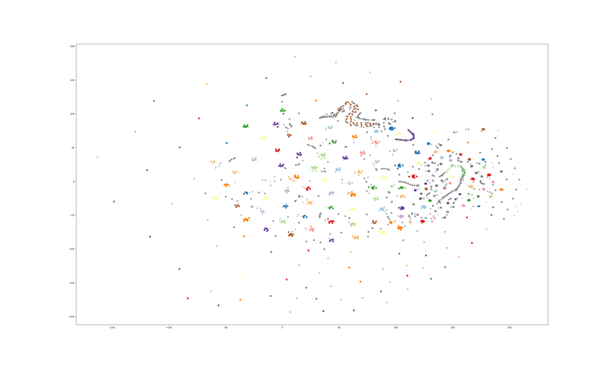}
			\captionsetup{justification=centering}
			\caption{t-SNE Graph With 110 Features Before Feature Selection}
			\label{fig:t-SNE_before}
		\end{subfigure}
		\hfill
		\begin{subfigure}[b]{\textwidth}
			\centering
			\includegraphics[width=\textwidth]{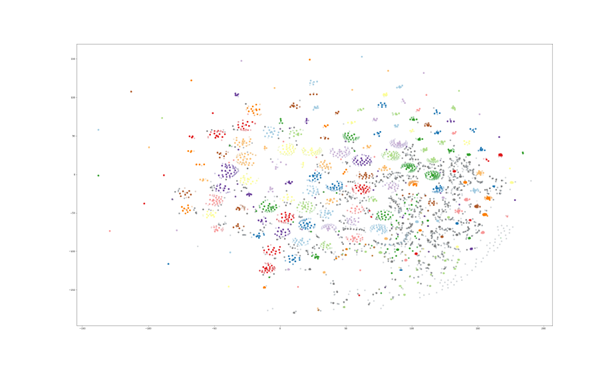}
			\captionsetup{justification=centering}
			\caption{t-SNE Graph With 12 Features After Feature Selection}
			\label{fig:t-SNE_after}
		\end{subfigure}
		\caption{t-SNE Graphs Before and After Feature Selection}
		\label{fig:t-SNE}
	\end{figure}

	The ANOVA test calculated the variance of each feature to return their significance scores. In combination with ablation results, the 12 features returned from the ANOVA test were selected. Adding features beyond the top 12 created more noise in the t-SNE graphs (See Figure \ref{fig:t-SNE}). The ANOVA test identified the following features as the most influential: history of a broken hip, history of a broken spine, history of a broken wrist, no history of a broken wrist, no history of a broken spine, daily prednisone or cortisone use, father has a history of a hip fracture, mother has a history of a hip fracture, mother was diagnosed with osteoporosis, father was diagnosed with osteoporosis, no history of a broken hip, and diagnosed with osteoporosis.
	
	HDBSCAN was performed iteratively over the dataset with all 12 features in each iteration. After running one iteration of HDBSCAN, the density of the clusters was calculated using a ``k-nearest neighbor search" on the mediods. The clusters produced by HDBSCAN may not be spherical, so a medoid better represents the cluster center than the centroids. The program calculated the five nearest points and their associated distances to return the densities of the clusters. I then sorted the clusters by density to find the labels of the densest clusters.
	
\begin{figure}[!h]
		\begin{subfigure}[b]{\textwidth}
			\centering
			\includegraphics[width=\textwidth]{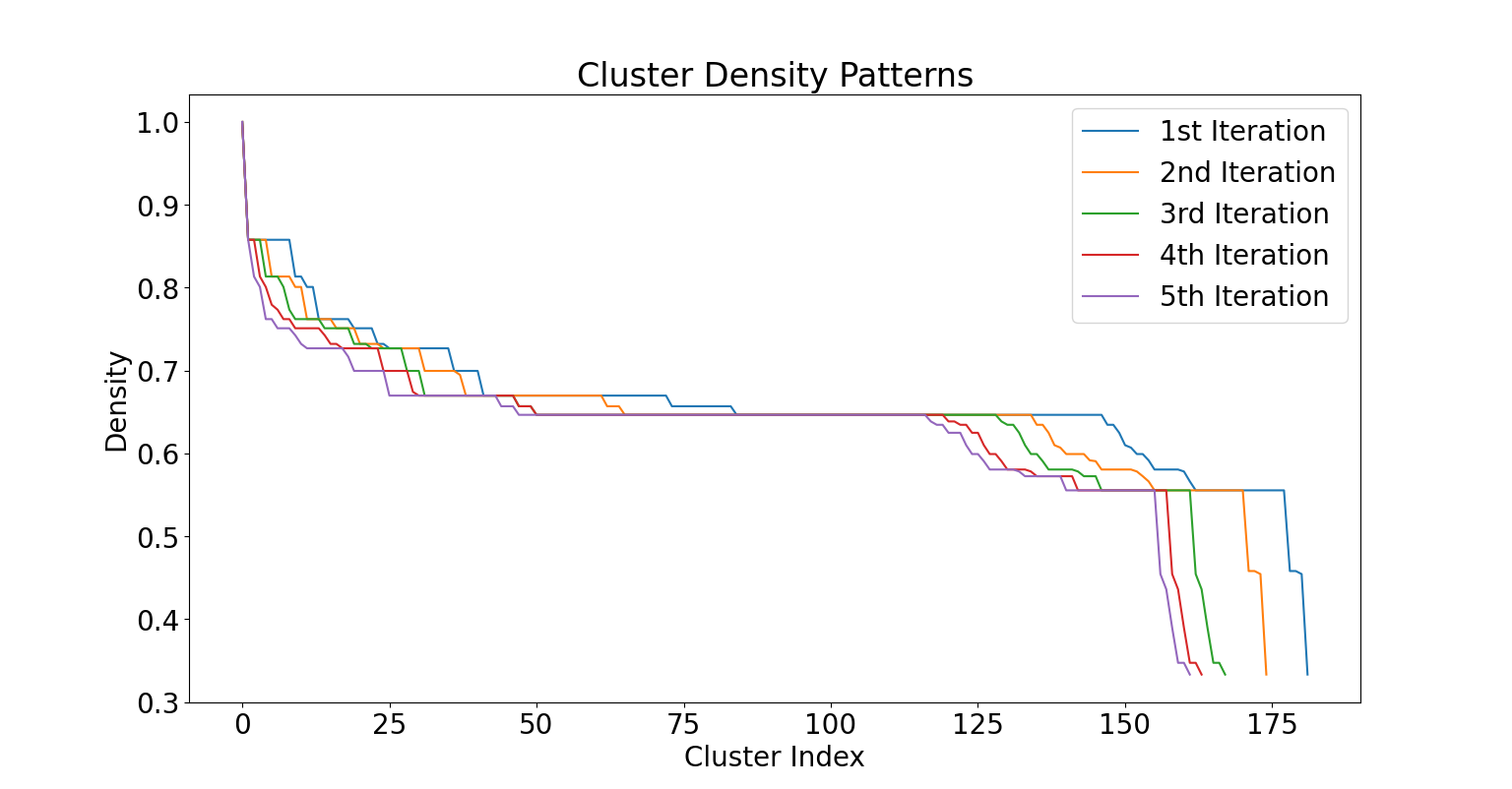}
			\captionsetup{justification=centering}
			\caption{Clusters Sorted by Density At 5 Iterations}
			\label{fig:clust_dens}
		\end{subfigure}
		\hfill
		\begin{subfigure}[b]{\textwidth}
			\centering
			\includegraphics[width=0.8\textwidth]{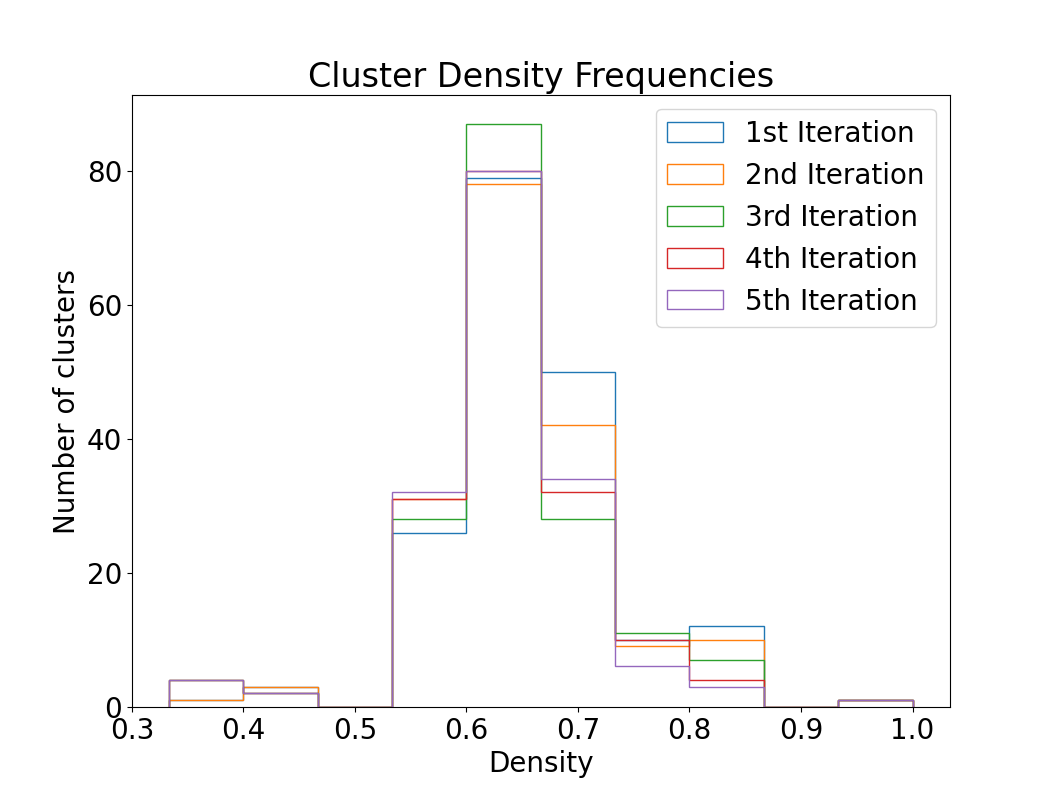}
			\captionsetup{justification=centering}
			\caption{Frequencies of the Cluster Densities At 5 Iterations}
			\label{fig:clust_freq}
		\end{subfigure}
		\caption{Cluster Density Graphs}
		\label{fig:clust_graph}
	\end{figure}

	I constructed a density pattern graph by sorting the clusters by density and plotting the drops and gaps in density on a graph. The density values range from 0.0 to 1.0, where 0.0 is the least dense and 1.0 is the most dense. I used 0.85 as the density threshold for determining dense clusters because I noticed that the differentiation in the number of clusters present for each iteration began at the 0.85 density mark on the graph (See Figure \ref{fig:clust_graph}). I removed the clusters 0.85 or greater in density and re-clustered the remaining data at the same threshold. I repeated this process for five iterations until the clustering algorithm returned two dense clusters. The number of clusters with a density of 0.85 or greater decreased with each iteration, which is typical in creating a hierarchy of clusters. Proceeding with a sixth iteration would not return any dense clusters to analyze. 
	
	The study also used the same iterations to analyze large, less dense clusters to potentially find patterns in sparse clusters. When selecting a lower bound for the cluster density, I used 0.65 because the cluster density pattern graph depicted a large plateau in which there were many clusters with this density, with a few larger clusters in that area of the graph. A large cluster was defined by a cluster size of at least 2,150 because it is twenty times as large as the average dense cluster size (See Figure \ref{fig:clust_graph}).
	
	Once the dense clusters are selected, the next step is identifying significant principal features that determine cluster groups and differentiate one cluster from another. The Wasserstein distance represented the dissimilarity in features between clusters, which would determine the significance of that feature. In this study, I considered a categorical feature as a principal feature if it returned a Wasserstein distance of 1.0 because the values could only be 0.0 and 1.0; a distance of 1.0 indicated that the features had opposing values, and needed to be moved entirely to match the other distribution. As a result, that feature would contribute to cluster discrimination. For the age, the only numerical feature included in the 12 significant features, a Wasserstein distance of nine or more identified age as a principal feature, as nine years was 10\% of the age range, 0 to 90 years old.
	
	\subsection{Experimental Results}
	HDBSCAN returned several clusters in each iteration for the dense and sparse subsets. In the first iteration, HDBSCAN identified nine dense clusters and eight large sparse clusters. The second iteration of HDBSCAN retrieved five dense clusters after removing the dense clusters in the first iteration, as well as five large sparse clusters. The third iteration of HDBSCAN found five dense clusters and two large sparse clusters. The fourth iteration of HDBSCAN identified three dense clusters and two sparse clusters and did not find any large sparse clusters. The fifth iteration of HDBSCAN returned two dense clusters and one sparse cluster.
	
\begin{figure}[!h]
		\centering
		\includegraphics[width=\linewidth]{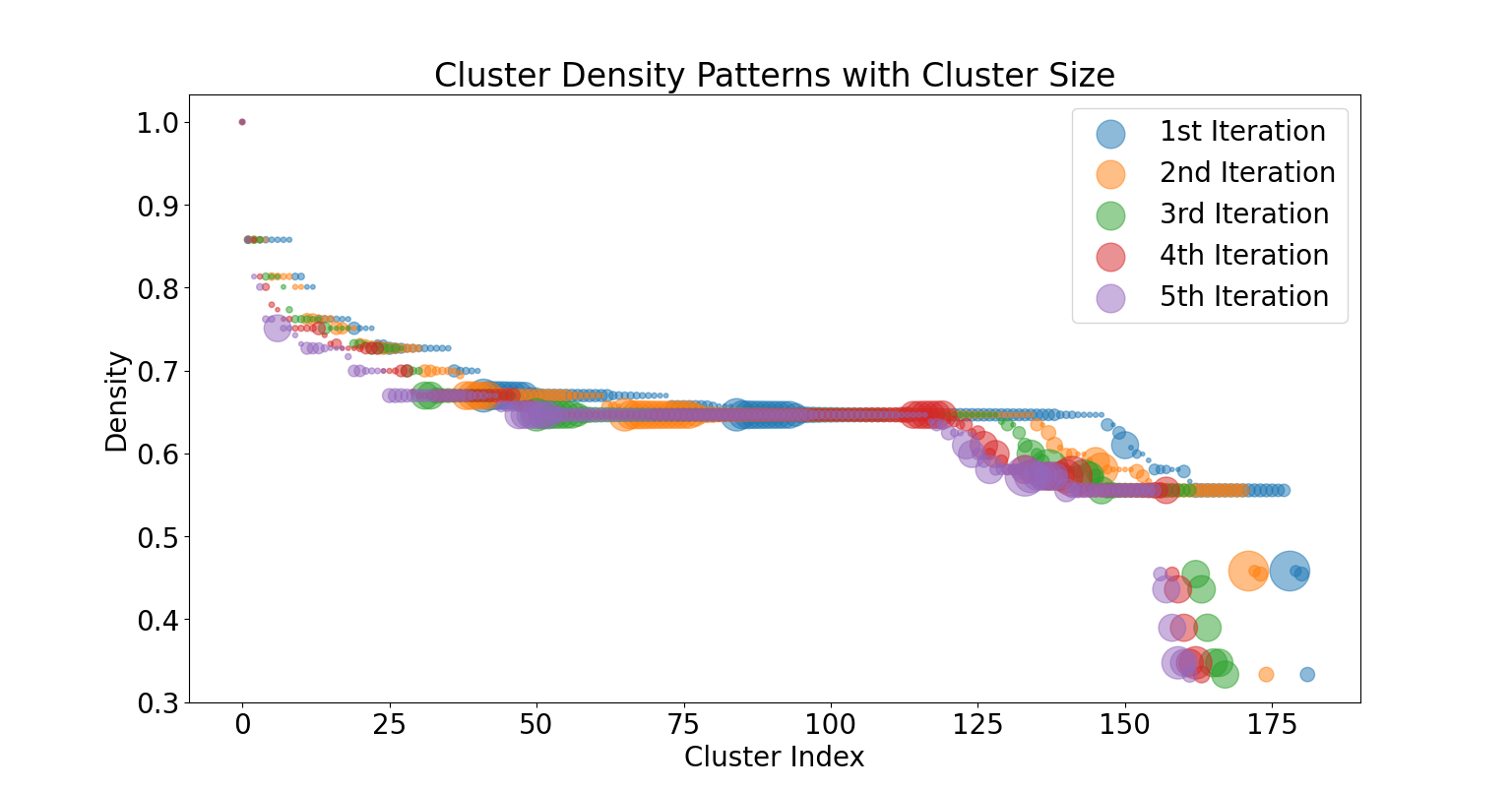}
		\captionsetup{justification=centering, margin=1cm}
		\caption{Graph of Cluster Sizes Sorted By Density}
		\label{fig:clust_size}
	\end{figure}

	The densities of the dense clusters at each iteration were 0.85 or greater in density, with cluster sizes ranging from 51 to 175 subjects in size. The average size of these dense clusters was 107.5 data points. The sparse clusters identified at each iteration had a density between 0.6699 and 0.7511. The average size of the large sparse clusters in all iterations was 2,463 data points (See Figure \ref{fig:clust_size}).
	
	Wasserstein distance identified multiple principal features of the dense clusters within each iteration. The first iteration of HDBSCAN returned age, history of a broken wrist, and no history of a broken hip, no history of a broken spine, and no history of a broken wrist as principal features. The second HDBSCAN iteration found that age, having a mother who was diagnosed with osteoporosis, having a father who was diagnosed with osteoporosis, a history of a broken wrist, no history of a broken hip, no history of a broken spine, and no history of a broken wrist are principal features. The third iteration identified the principal features to be age, daily prednisone or cortisone use, history of a broken wrist, no history of a broken hip, no history of a broken spine, and no history of a broken wrist. The fourth iteration of HDBSCAN discriminated clusters based on age, daily prednisone or cortisone use, having a mother who was diagnosed with osteoporosis, and having a father who was diagnosed with osteoporosis. The fifth iteration of the HDBSCAN separated dense clusters based on age, history of a broken wrist, no history of a broken hip, and no history of a broken spine. Age was the only principal feature found amongst the large and sparse clusters in each iteration.
	
	\paragraph{Conclusions} The results of this study undermine previous studies by demonstrating many counter examples to the identified risk factors of osteoporosis, thus challenging the findings of existing research. In previous works, research established increasing age, female sex, White race, early ovary removal, prolonged corticosteroid use, prior fractures, and family history of osteoporosis as risk factors for osteoporosis. This study identified age, daily corticosteroid use, history of previous fractures, and family history as principal features of dense clusters, consistent with other research. However, the cluster sizes of these groups were small, averaging 107.5 data points each, compared to the 101,316 subjects included in the study. The small cluster size suggests that of the participants in the NHANES dataset, only a few exhibited the risk factors presented in previous research. Additionally, the ANOVA test did not identify gender, race, and ovary removal as significant factors in determining osteoporosis risk levels, which is inconsistent with previous research that found these features noteworthy. These results suggest that previous research may be unreliable and that more research is needed to investigate these identified risk factors.
	
	\section{Discussion}
	The study has limitations in which further research is needed to collect and analyze more detailed information to obtain significant results regarding osteoporosis risk factors. Potential studies include lowering the density threshold of the clusters, studying sparse clusters with larger sample sizes, removing missing data due to dependencies on other collected data, and analyzing the resulting cluster shapes. This study used 0.85 for the density threshold when determining dense clusters; however, the threshold resulted in small clusters that were not significant enough for a dataset containing 101,316 subjects. Perhaps a lower threshold would result in larger clusters in subsequent iterations and present substantial results. Another study can use the 0.85 density threshold and associated iterations used in this study but focus on the clusters located between the 0.55 to 0.65 densities as the density pattern graph shows larger and more differentiated clusters. There is potential in finding results in the sparse clusters, unlike the dense clusters used in this study. Another problem present in this study is the amount of missing data in the database, often due to nonapplicable follow-up questions. This study hoped to find patterns in the missing data; however, the large amount of imputed data may have altered the results. I would recommend that further studies focus on the data containing participant answers to the follow-up questions and have minimal missing data. Last, this study uses HDBSCAN clustering because the algorithm can detect clusters of abnormal shapes that are not spherical or ball-shaped. The shape of the NHANES data clusters is unknown, and it may be beneficial to conduct a study analyzing the shape of the resulting clusters using HDBSCAN.
	
	\section{Conclusion}
	This project studied the reliability of identified risk factors associated with osteoporosis. One study established age, gender, ethnicity, removal of the ovaries at an early age, and prolonged use of corticosteroids as risk factors of osteoporosis \cite{kelsey1989risk, kanis1998risk, pouresmaeili2018comprehensive}. Another study concluded that risk factors included history of osteoporosis, number of pregnancies, diabetes, and diabetes-related medications \cite{jabarpour2020osteoporosis, kanis1998risk, pouresmaeili2018comprehensive}. This study proposes utilizing the CLIF to identify dense clusters and sub-clusters in a data set and the Wasserstein distance to identify principal features in the selected clusters. The study also proposes using a combination of ANOVA and ablation tests to select influential features from a data set. These proposed methods found multiple dense clusters with a density of 0.85 or greater in five iterations of HDBSCAN. These clusters were separated by the following principal features: age, history of a wrist fracture, no history of a wrist fracture, no history of hip fracture, no history of a spinal fracture, daily prednisone or cortisone use, mother was diagnosed with osteoporosis, and father was diagnosed with osteoporosis. The features of the dense clusters are consistent with previous research; however, the small cluster sizes undermine prior studies by demonstrating many counter examples to the identified risk factors of osteoporosis. This conclusion suggests that previous studies may be unreliable regarding the identified risk factors, and further studies are needed to better understand these correlational relationships.
	
	\section{Future Works}
	The key contributions of this study include the implementation of the CLIF and principal feature identification with Wasserstein. The methodology of this study utilized iterative clustering with HDBSCAN to find subtle differences in the data clusters and subclusters. Iterative clustering works beyond HDBSCAN, and it is a framework that applies to any clustering algorithm. Principal feature identification using Wasserstein distance metrics is another key contribution. Wasserstein distance can capture underlying structures and nuances of probability distributions, making it a valuable tool for identifying principal features that discriminate one cluster from another. Wasserstein distance normalizes data before calculating the distance, allowing distributions to be of different sizes. With its distribution normalization, Wasserstein can be implemented in other studies regardless of whether distributions have the same size. Future studies may implement iterative clustering and Wasserstein distance metrics to find hidden patterns in other data and use similar methods to study the reliability of risk factors in other diseases.
	
	\section{Acknowledgments}
	This work would not have been possible without academic rigor and the supportive faculty of Misericordia University. I am especially indebted to Dr. Fanchao Meng, my Capstone advisor, and professor of several math and computer science courses; Sr. Patricia Lapczynski, director of the computer science program, professor to many of my computer science classes, and my academic advisor; and Dr. Joseph Curran, director of the Honors program, who have been supportive throughout the Capstone project process and who worked to provide the time and guidance needed to complete the project. I would also like to thank Sr. Patricia Lapczynski; Dr. Fanchao Meng; Dr. Steven Tedford, professor and chair department of mathematics and data science program director; Dr. John Woznicki, Dean of Arts and Sciences; Ms. Michelle Donato, administrative assistant to the President; and Mr. David Johndrow Jr., director of information technology, for their support in funding the resources and technology required for the research project's execution and success.
	
	\newpage
	\bibliographystyle{plain}
	\bibliography{osteoporosisRisk}
\end{document}